\documentclass[conference]{IEEEtran}
\usepackage{cite}
\usepackage{graphicx}
\usepackage[utf8]{inputenc}
\usepackage[english]{babel}
\usepackage{multicol}
\usepackage{mathtools}
\usepackage{url}
\DeclarePairedDelimiter{\abs}{\lvert}{\rvert}

\graphicspath{ {images/} }

\usepackage{array,booktabs,arydshln,xcolor}

\def\BibTeX{{\rm B\kern-.05em{\sc i\kern-.025em b}\kern-.08em
    T\kern-.1667em\lower.7ex\hbox{E}\kern-.125emX}}
\begin{document}

\title{Table Structure Extraction with Bi-directional \\
Gated Recurrent Unit Networks}


\author{\IEEEauthorblockN{Saqib Ali Khan${ }^{1}$, Syed Muhammad Daniyal Khalid${ }^{1}$, Muhammad Ali Shahzad${ }^{1,2}$ and Faisal Shafait${ }^{1,2}$}
\IEEEauthorblockA{${ }^{1}$ School of Electrical Engineering and Computer Science (SEECS),\\
		National University of Sciences and Technology (NUST), 
		Islamabad, Pakistan\\
		${ }^{2}$ Deep Learning Laboratory, National Center of Artificial Intelligence (NCAI), Islamabad, Pakistan\\
		Email: faisal.shafait@seecs.edu.pk}}

\maketitle

\begin{abstract}
Tables present summarized and structured information to the reader, which makes table’s structure extraction an important part of document understanding applications. However, table structure identification is a hard problem not only because of the large variation in the table layouts and styles, but also owing to the variations in the page layouts and the noise contamination levels. A lot of research has been done to identify table structure, most of which is based on applying heuristics with the aid of optical character recognition (OCR) to hand pick layout features of the tables. These methods fail to generalize well because of the variations in the table layouts and the errors generated by OCR. In this paper, we have proposed a robust deep learning based approach to extract rows and columns from a detected table in document images with a high precision. In the proposed solution, the table images are first pre-processed and then fed to a bi-directional Recurrent Neural Network with Gated Recurrent Units (GRU) followed by a fully-connected layer with softmax activation. The network scans the images from top-to-bottom as well as left-to-right and classifies each input as either a row-separator or a column-separator. We have benchmarked our system on publicly available UNLV as well as ICDAR 2013 datasets on which it outperformed the state-of-the-art table structure extraction systems by a significant margin.

\end{abstract}


\IEEEpeerreviewmaketitle

\section{Introduction}
A table contains an ordered arrangement of rows and columns that are widely used to present a set of facts about some information~\cite{lewandowsky}. They are widely used in research articles, data analysis, newspapers, magazines, invoices and financial documents. Tables present multiple information points for a large number of items in rows and columns that are easy to perceive and analyze. They structure the information to provide a visual summary of the most valuable information contained in the document. It is for this reason that the table recognition systems have captured the interest of a large number of researchers to make contributions in this domain over the past two decades.

Tables have numerous layouts which makes it very hard for conventional feature engineering approaches to decode table structures generically. These approaches generally rely on visual features like ruling lines, spacing between different columns, type of data in the table cells, their relationships with overlapping neighbors or color encoded cell blocks. They perform reasonably well on the tables of a particular layout or a business case but fail to scale across multiple domains.

In the recent years, researchers have greatly improved the results of computer vision problems by applying deep learning techniques. Schreiber et al.~\cite{schreiber_icdar17} proposed a deep learning based approach for recognizing rows and columns of tables in document images. Their proposed system employs a semantic segmentation (FCN-Xs architectures) model with custom tweaking to the hyper-features as well as skip pooling to enhance the segmentation results. The major limitation of this method is the way FCN processes the table. Each stride of an FCN filter maps a portion of the input image pixels to an output pixel. This fails to capture the fact that the rows and columns in a table follow a unique repetitive sequence of in-between spacing and data length as the information of the next and the previous row-column elements is not taken into account. Also, the receptive field of the CNN based models does not process the entire row or column in a single stride. In this paper, we overcome this limitation by using a sequential modeling approach. Specifically two bi-directional GRUs are used. One bi-directional GRU identifies the row boundaries while the other identifies the column boundaries. Each bi-direction GRU has its own fully connected layer to classify the input as either a row-boundary or a column-boundary. Our approach successfully overcomes the limitations of a CNN based model and provides a data-driven approach towards a general, layout independent table structure extraction system. 
We have benchmarked our system on publicly available UNLV dataset~\cite{unlv} where it outperformed T-Recs~\cite{kieninger_das98, kieninger_icdar01} table structure recognition system. It is to be noted that no part of the UNLV dataset has been used in the training process.

The rest of this paper is organized in the following sections: Section II consists of the related work in table structure recognition domain. Section III elaborates our proposed methodology that consists of a pre-processing module and a classification module. Section IV presents the evaluation metrics while benchmarking and evaluation of the proposed algorithm is detailed in Section V. Section VI provides the conclusive remarks as a guideline for the future work in this domain.

\section{Related Work}
A substantial amount of work has been done to identify the structure of a table both using heuristic-based methods as well as using deep learning. Kieninger et al.~\cite{kieninger99,kieninger_das98,kiegner_icapr99} proposed a system which was one of the earliest successful attempts on table structure extraction problem called {T-Recs}. The input to this system is the word bounding boxes. These boxes are then grouped into rows and columns using a bottom-up approach by evaluating the vertical and horizontal overlaps between the boxes to form a segmentation graph. The major problem in this approach is that the output depends on a large number of parameters values that are heuristically set. Besides, the algorithm fails if the preceding OCR step does not correctly identify words bounding boxes (for example if the character recognizer misses dots and commas in numeric data).

Wang et al.~\cite{wang_04} proposed a data-driven approach similar to the X-Y cut algorithm~\cite{shafait_pami08} that is based on probability optimization technique to solve table structure extraction problem. This statistical algorithm uses probabilities that are derived from a large training corpus. This method also takes into account the distances between adjacent words and it works on single column, double column and mixed column layouts.

Shigarov et al.~\cite{shigarov_16} proposed a method that relies on PDF metadata with information including font and text bounding boxes. The algorithm uses ad-hoc heuristics for recovering table cells from text chunks and ruling lines. The algorithm combines these text chunks into text blocks through a text block recovery algorithm and then uses a threshold to configure the block vertically or horizontally.

Zanibbi et al.~\cite{zanibbi_das_04} presented a survey for table recognition systems in terms of interactions of table models, observations, transformations, and inferences. Their survey answers questions about what and when some decisions are made by table structure recognition systems. Furthermore, this survey outlines the dataset used for the training and evaluation of these systems.

Jianying et al.~\cite{jianying01} proposed a general algorithm for table structure extraction from an already detected table region. In their proposed methodology, they have used hierarchical clustering for column detection. Additionally, the system uses lexical and spatial criteria to classify headers of tables. They have used a directed acyclic attribute graph or DAG for evaluation of table structure extraction.

Wang et al.~\cite{wangt_dar01} proposed an automatic ground truth generation system which can generate a large amount of accurate ground truth for table recognition systems. They use novel background analysis table recognition algorithms and an X-Y cut algorithm for identifying table regions in the document images. This system takes line and word segmentation results as input and outputs table cell detection results.

Kasar et al.~\cite{Kasar_icdar_15} proposed a technique for table structure extraction based on query-patterns. This approach is a client-driven approach in which the client will provide the query pattern based on the location of key fields in the document. The input query pattern is then converted into a relational graph in which the nodes represent the features and the edges represent the spatial relationship between these nodes.

Shamilian et al.~\cite{shamilian} proposed a system that reads layout of the tables in machine printed form. They have provided a graphical user interface (GUI) for users to define contextual rules to identify key fields inside a table. The system can also be manually retargeted to new layouts by the user. This system has been applied to more than 400 distinct tabular layouts.

Schreiber et al.~\cite{schreiber_icdar17} proposed a deep learning based approach for table structure recognition. This system uses semantic segmentation model with FCN-8 architecture and skip pooling features to detect rows and columns of a table. Additionally, they have vertically stretched the table images in order to increase the precision on row detection. Furthermore, they have used CRF to improve the results of semantic segmentation model. Siddiqui el al.~\cite{siddiqui_decnt} also proposed a deep learning based method based on Deep Deformable Convolutional Neural Network (CNN) for table detection.

In this paper, we have proposed a novel solution for table structure extraction using a sequential model, assuming that the table has already been detected using an existing algorithm (e.g.~\cite{gilani_icdar17}. In the proposed methodology, the table images are first pre-processed by applying binarization, noise removal, and morphological transformation. These transformed images are then passed to a bi-directional Gated Recurrent Unit (GRU) recurrent neural network that detects rows and columns in the table.

\section{Proposed Methodology}
The proposed method is divided into three modules: Image pre-processing, a row-column classifier and post-processing. The pre-processing step plays a crucial role in converting the table images containing text to natural images that do not contain textual features. These images are then passed to the classifier that uses rows and columns as time steps to classify each row and column. In the post-processing step, the segmentation space generated by the classifier is parsed to give a single line prediction of rows and columns. This section explains each module in greater detail.

\subsection{Image Pre-processing}
The first and the foremost step is pre-processing the table images. This step plays a preliminary role in converting the raw table images to a simpler form so that the layout or structure of the table is more apparent. The goal of this transformation is to increase the efficiency of our classifier by removing unnecessary detail from the input images. 

The images are first cleaned up by removing the ruling lines and other non-text foreground objects. The cleaned image is then run through adaptive binarization~\cite{shafait_binarize08} so that the pixel intensities are uniform. Once the images have been binarized, they are resized to a fixed dimension of $1600 \times 512$ as the neural network is designed to process fixed size inputs.

\begin{figure}[ht]
\centering
\includegraphics[scale = 0.190]{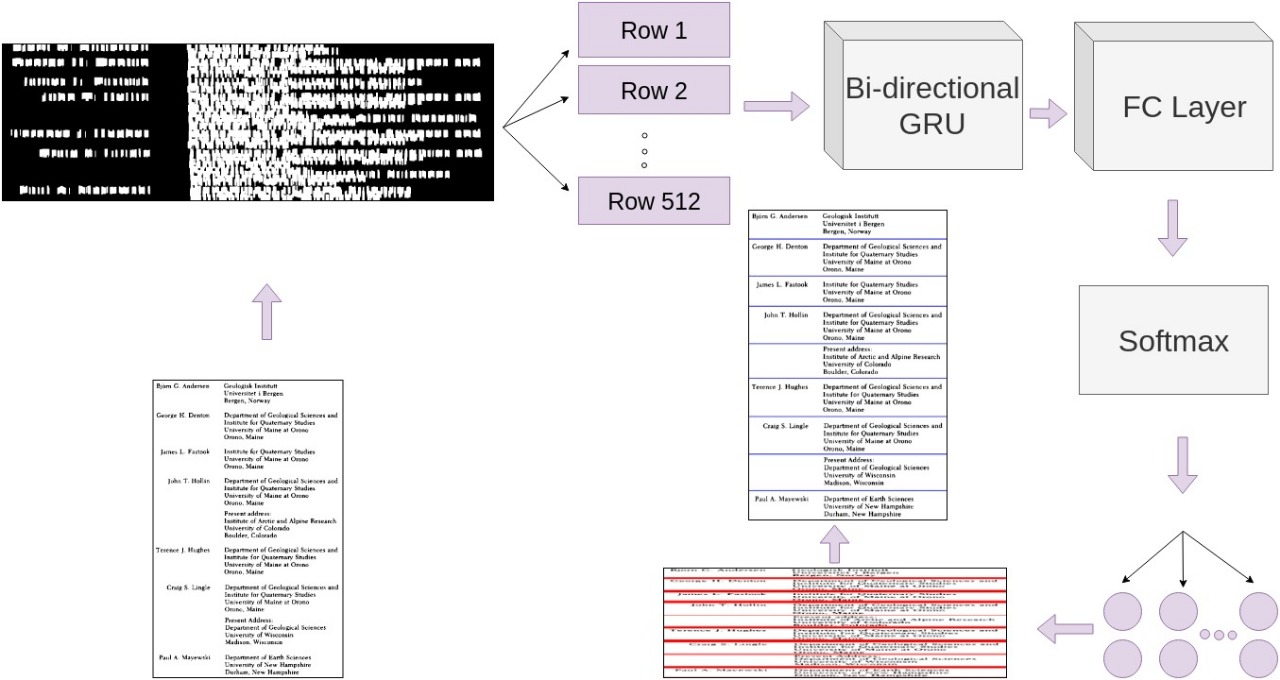}
\caption{Neural Architecture for row classification: Passing a $(1600 \times 512)$ pre-processed image to a bi-directional GRU with an input of size $(1600 \times 1)$ at each timestep. The bi-directional GRU outputs a $(512 \times 2)$ vector which is post-processed to get a single regressed row segmentation boundary.}
\label{row}
\end{figure}

After binarization, three iterations of dilation transform are applied to the resized image using a rectangular kernel. In the case of column detection, the dilation kernel is a vertical dilation filter of dimensions $3 \times 5$ and in the case of row detection, it is a horizontal dilation filter of dimensions $5 \times 3$. These dilation operations join the adjacent rows and columns, which helps the model to pick up the pattern of the row and the column separators. The transformed images are then normalized to have values between 0 and 1 to be fed to the subsequent recurrent neural network.

\subsection{Model}
This section provides details of the proposed methodology and it is further divided into two parts i) Column Classification ii) Row Classification. These two tasks are not very different by nature yet they require different model organization.

The crux of our approach is to identify segmentation space between the rows and the columns using recurrent neural networks. Different architectures of recurrent neural networks are proposed in the literature. We have selected Gated Recurrent Unit (GRU)~\cite{schuster, chung14} and Long Short-Term Memory (LSTM) networks~\cite{Kasar_icdar_15, shamilian} for our algorithm because of their ability to incorporate contextual information without vanishing gradient problem. The results demonstrate (see Section~\ref{results} that GRUs outperform the LSTMs by a significant margin for both row and column classification. An analysis of the results showed that the LSTM networks, due to their inherent complexity, tend to overfit on the simpler data. The later sections in this paper will only discuss the approach with GRUs for brevity, as the approach with LSTMs is quite similar.

The bi-directional GRU takes rows and columns as timesteps and use the information of previous row-column elements to predict future ones. This approach provides a significant improvement over the CNN based models because of the memory cells in GRUs that learns the pattern of inter-row and inter-column spacing and the sequence of repetition of row-column elements effectively. This approach outperformed Schreiber et al.~\cite{schreiber_icdar17} table structure extraction system based on semantic segmentation by a significant margin. The architectures for row and column classification are detailed in the following two sections.

\begin{figure}[ht]
\centering
\includegraphics[scale = 0.190]{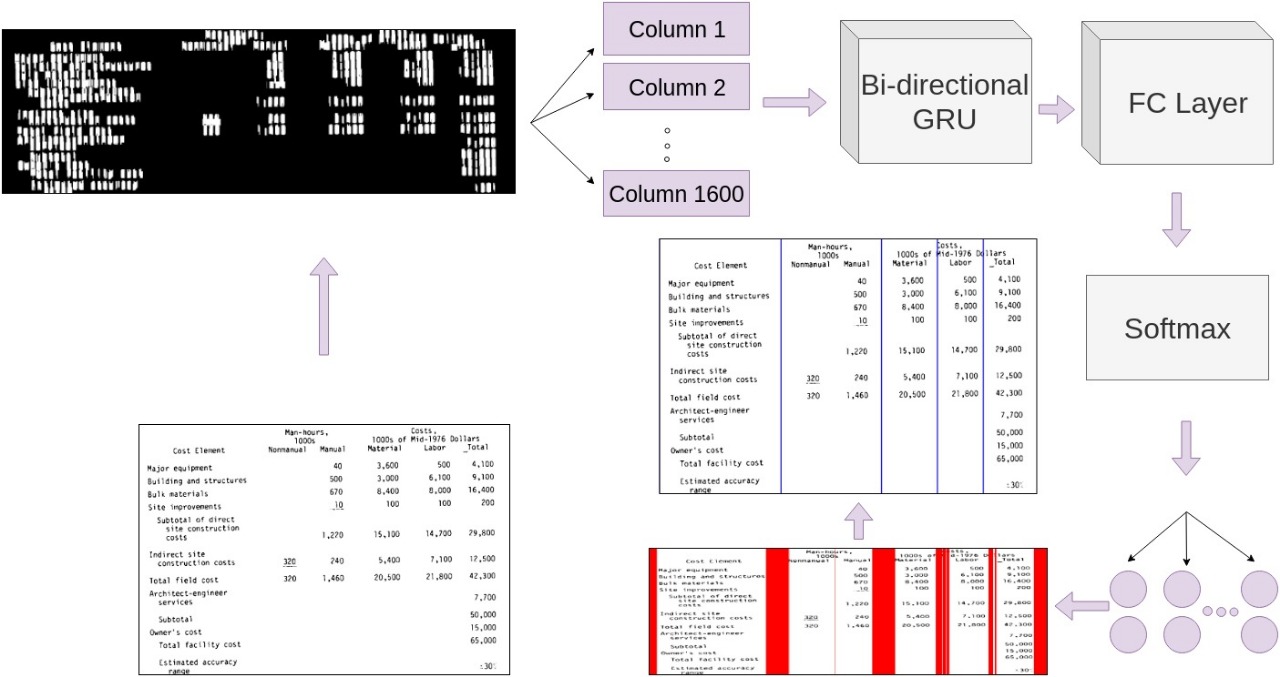}
\caption{Neural Architecture for column classification: Passing a $(1600 \times 512)$ pre-processed image to a bi-directional GRU with an input of size $(512 \times 1)$ at each timestep. The bi-directional GRU outputs a $(1600 \times 2)$ vector which is post-processed to get a single regressed column segmentation boundary.}

\label{column}
\end{figure}

\subsubsection{Column Classification}
The neural architecture for column recognition classifies each column of the image as either a column or a whitespace between two columns. The images are passed one at a time and each image is considered to be a batch like in stochastic gradient descent (SGD). The pre-processed input image of dimension $1600 \times 512$ within a single batch is split into $1600$ sequences (columns), each consisting of $512$ pixel values. We have used a hidden dimension of size $512$. The two-layer GRU is initialized with hidden dimensions $(4 \times 1 \times 512)$ corresponding to 2 * number of layers * batch size * hidden dimension size.

The GRU processes the image as $1600$ “timesteps”, each timestep corresponding to a column with $512$ input pixel values. At each timestep, the GRU has the information about all the columns to the left and the right (if any) of the current column, as well as the pixel values contained within the current column being evaluated. Using this information, the GRU can learn to identify the gap between the columns as those columns containing mostly white pixels and having two column regions on their left and right sides.

The GRU outputs a tensor of shape $1600 \times 512$ corresponding to sequence length times hidden dimension. This tensor is then passed through a fully connected layer which outputs a $1600 \times 2$ shaped tensor. This output is finally passed through a softmax layer which gives the final output of shape $1600 \times 2$, consisting of binary class probabilities for each of the $1600$ columns.

\subsubsection{Row Classification}
The neural architecture for row detection is a transpose of the column classifier and it classifies each row of the image as either a row or a whitespace between two rows. The images are fed one at a time and each image is considered to be a batch. The pre-processed input image of dimension $1600 \times 512$ within a single batch is split into $512$ sequences (rows), each consisting of $1600$ pixel values. We have used a hidden dimension of size $1024$. The 2-layer GRU is initialized with hidden dimensions $(4 \times 1 \times 1024)$ corresponding to 2 * number of (layers * batch size * hidden dimension size).

In the case of row classification, there are $512$ timesteps with $1600$ inputs each. At each timestep, the GRU has information about all the rows above and below the current row as well as the pixel values within the current row.

The GRU outputs a tensor of shape $512 \times 1600$ corresponding to sequence length x hidden dimension. This tensor is then passed through a fully connected layer which outputs a $512 \times 2$ shaped tensor. This output is finally passed through a softmax layer which gives an output of shape $512 \times 2$, consisting of binary class probabilities for each of the $512$ rows.

The last step in the classification is parsing the segmentation space predicted by the classifier. We take the midpoint of the segmentation space and applying the logic to drop the leftmost and the rightmost predictions in the case of columns and the top and the bottom predictions in the case of rows. This step regresses the output to a single line prediction of rows and column separators.

The complete model architectures for row and column classification are shown in Figure~\ref{column} and~\ref{row}.

\subsection{Training}
We used Adam optimizer paired with binary cross entropy loss function to train our models. A typical table image contains more rows and columns than the whitespace between them. Initial attempts at training resulted in a model that always predicted a row-column element and failed to detect the whitespace due to this class imbalance problem. So, we took measures to reduce this class imbalance problem and applied weighting to our loss function to penalize an incorrectly predicted row-column element only $66\%$ as much as an incorrectly predicted whitespace element.

The dataset used for training consisted of freely available document images downloaded from various sources. The tables, rows and columns were manually labelled using custom tools. With a fixed learning rate of $0.0005$, we trained the column classifier for $10$ iterations over $323$ images and the row classifier for $35$ iterations for $286$ images.

\section{Performance Measures}
Various researchers have used different evaluation metrics ranging from simple precision and recall to more complex evaluation algorithms. In this paper, we have used the performance evaluation algorithm described in Shahab et al.~\cite{shahab_das10} to evaluate the performance of our model for two main reasons: i) This metric paints the detailed picture of how the algorithm performs using six different measures. ii) It is a general purpose metric that can be applied to any type of segments such as tables, rows, columns and cells.

The methodology proposed by Shahab et al.~\cite{shahab_das10}, starts with numbering the ground truth segments and the detected segments. A correspondence matrix is then created with $m$ rows and $n$ columns where $m$ is the number of ground truth segments and $n$ is the number of detected segments in an image. The [G\textsubscript{i},S\textsubscript{j}] entry in the matrix represents the number of pixels in the $ i^{th} $ ground truth segment that overlap with the $ j^{th} $ detected segment. $ \abs{G_i} $ represents the total number of pixels in the $i^{th}$ ground truth segment and  $\abs{  {S_j}  }$   represents the total number of pixels in the $ j^{th} $ detection. Once the correspondence matrix for an image has been created, we can define the following measures:

\begin{figure*}[ht]
\centering
\includegraphics[scale = 0.35]{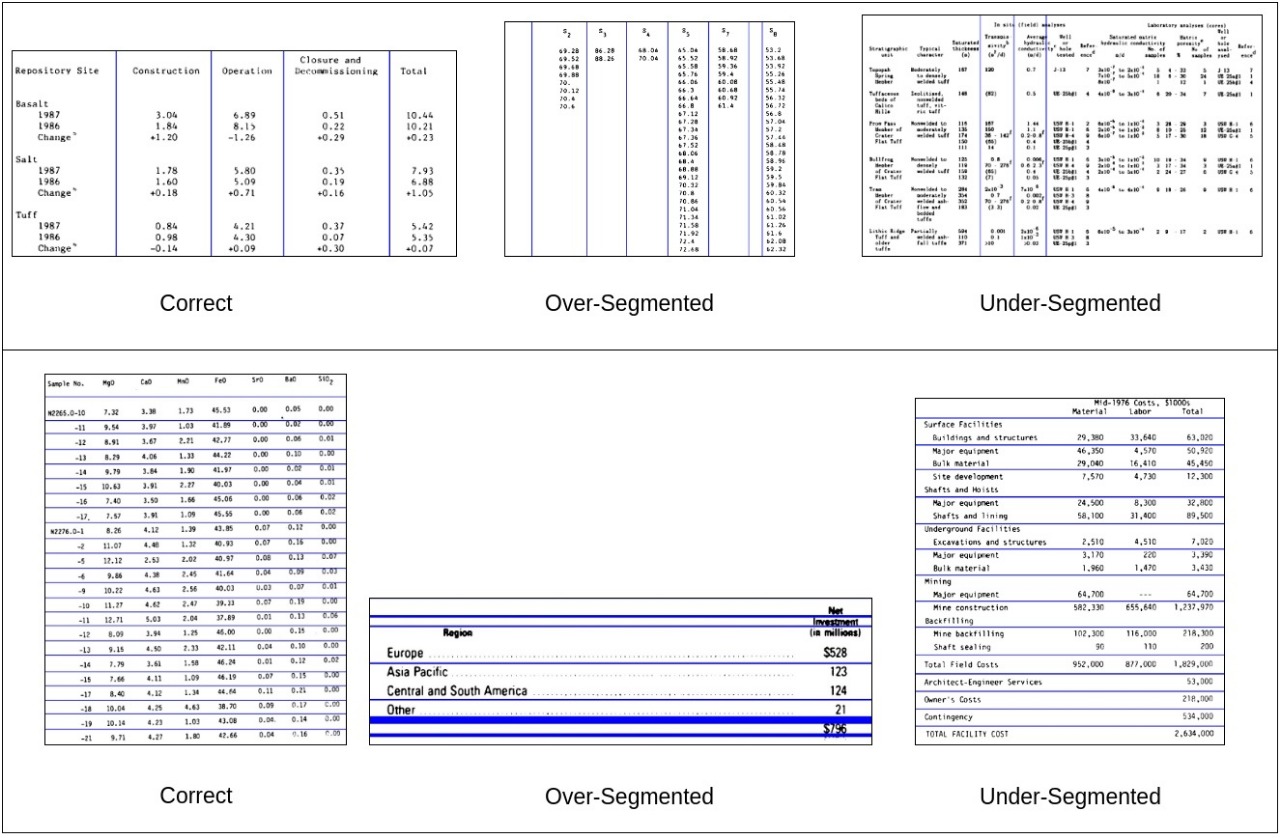}
\caption{Results of our proposed table structure extraction approach on a few images from UNLV dataset showing examples of correct, over-segmented and under-segmented detections}
\label{results}
\end{figure*}

\subsection{Correct Detections}
This measure shows the total number of ground truth segments that have a large intersection with a detected segment and the detected segment does not have a significant overlap with any other ground truth segment. That is, for a detected segment $S_j$ and ground truth segments $G_i$:

\vspace{1pt}

\begin{center}
$  \frac{\abs{G_i \cap S_j}}{\abs{G_i}} > 0.9  $ and $ \frac{\abs{G_k \cap S_j}}{\abs{S_j}} < 0.1$ \hspace{4pt}    $    \forall k \neq i  $
\end{center}

\subsection{Partial Detections}
The total number of ground truth segments that have a significant intersection with a single detected segment. However, the intersection is not large enough to be counted as a correct detection. That is, 

\vspace{1pt}
\begin{center}
$ 0.1 < \frac{\abs{G_i \cap S_j}}{\abs{G_i}} < 0.9 $ and $ \frac{\abs{G_i \cap S_k}}{\abs{G_i}} < 0.1$ \hspace{4pt}    $    \forall k \neq j  $
\end{center}

\subsection{Over segmentation}
The total number of ground truth segments that have a large overlap with two or more detected segments. An over-segmented detection means that multiple detected segments span over a single ground truth. Mathematically,

\vspace{1pt}

\begin{center}
$ 0.1 < \frac{\abs{G_i \cap S_j}}{\abs{G_i}} < 0.9 $ \hspace{4pt}  $ $ 
\end{center}

holds for more than one detected segments $S_j$ for a particular $G_i$.

\subsection{Under segmentation}
This is the inverse of over-segmentation i.e. the number of detected segments that have a large intersection with more than one ground truth segment. An under-segmented detection means that a single detection spans over multiple ground truth segments. Mathematically,

\begin{center}
$ 0.1 < \frac{\abs{G_i \cap S_j}}{\abs{S_j}} < 0.9 $ \hspace{4pt}  $  $
\end{center}

holds for more than one ground truth segment $G_i$ for a particular $S_j$

\subsection{Missed segments}
These are the number of ground truth segments that do not have a large overlap with any of the detected segments. These are segments that our algorithm should have detected but failed to do so.
\begin{center}
$ \frac{\abs{G_i \cap S_j}}{\abs{G_i}} < 0.1 $ \hspace{4pt}  $  \forall j  $
\end{center}

\subsection{False Positive Detections}
The inverse of missed segments, these are segments detected by the algorithm but which are not actually present in the ground truth. These are foreground pixels that algorithm has mistakenly detected as segments.

\begin{center}
$ \frac{\abs{G_i \cap S_j}}{\abs{S_j}} < 0.1 $ \hspace{4pt}  $  \forall i  $
\end{center}

Figure~\ref{results}  show our model output for some sample images on UNLV dataset showing correct, over-segmented and under-segmented detections.

\section{Experiments and Results}
We have used publicly available UNLV dataset~\cite{unlv} for the evaluation of our approach. The dataset spans a number of documents of varying layouts and domains including newspapers, research articles, magazines, technical reports etc. There are $2,889$ images in UNLV dataset~\cite{unlv}, out of which there are $427$ images containing at least one table. The dataset also includes accompanying, manually drawn ground truths for table boundaries. The ground-truth for columns, rows and cells was presented in~\cite{shahab_das10}. Since our work is focused on table structure extraction rather than table detection, we cropped the tables from the images using UNLV ground truth files, resulting in $557$ tables. We then evaluated our model’s outputs against the ground truths provided in~\cite{shahab_das10}. We did not use any image from the UNLV dataset in the training and validation process of our model and thus all the images were unseen by the model.

We have benchmarked our approach with {T-Recs} system by Kieninger et al.~\cite{kieninger99}, a non-deep-learning based algorithm for table structure extraction. Our approach provided a significant improvement in correct column detections, from $40.51\%$ to $55.31\%$ and from $54.98\%$ to $58.45\%$ in the case of row detection. On the other hand, the number of partial detections has gone down which is explained by the higher number of over-segmentations and under-segmentations in our approach as compared to {T-Recs} (see Table~\ref{table:column} and~\ref{table:row} for comparison results).

\begin{table*}
\centering
\caption{The results of evaluating our system on 427 binary 300-dpi scanned UNLV dataset pages containing table zones. The following benchmark is for column segmentation.}
\label{table:column}
\begin{tabular}{| l | r | r | r |}
    \hline
    & \multicolumn{3}{c}{\textbf{Accuracy\%}} \vline \\\cline{2-4}
    & & \multicolumn{2}{c}{\textbf{Our Approach}} \vline \\\cline{3-4}
    
    \textbf{Performance Measures} & \textbf{T-Recs} & \textbf{Bi-directional LSTM} & \textbf{Bi-directional GRU} \\ \hline
    
    Correct Detections & 40.51 & 49.05 & 55.31 \\ \hline
    Partial Detections & 18.57  & 15.13 &  12.13 \\ \hline
    Missed Detections & 13.50 & 6.99 & 3.12  \\\hline
    Over Segmented Detections & 13.50 & 18.44  & 12.14  \\ \hline
    Under Segmented Detections & 5.11 & 20.55  & 16.75 \\ \hline
    False Positive Detections & 0.88  & 1.20 & 0.08  \\
    \hline
    \end{tabular}
\end{table*}

\begin{table*}
\centering
\caption{Results of evaluating our system on 427 binary 300-dpi scanned UNLV dataset pages containing table zones. The following benchmark is for row segmentation.}
\label{table:row}
\begin{tabular}{| l | r | r | r |}
    \hline
    & \multicolumn{3}{c}{\textbf{Accuracy\%}} \vline \\\cline{2-4}
    & & \multicolumn{2}{c}{\textbf{Our Approach}} \vline \\\cline{3-4}
    
    \textbf{Performance Measures} & \textbf{T-Recs} & \textbf{Bi-directional LSTM} & \textbf{Bi-directional GRU} \\ \hline
    
    Correct Detections & 54.98 & 51.62 & 58.45 \\ \hline
    Partial Detections & 12.45  & 17.13 &  13.35 \\ \hline
    Missed Detections & 10.69 & 8.39 & 2.50  \\\hline
    Over Segmented Detections & 6.27 & 4.24  & 8.33   \\ \hline
    Under Segmented Detections & 7.70 & 5.30  & 14.67\ \\ \hline
    False Positive Detections & 0.12  & 0.59 & 0.15  \\
    \hline
    \end{tabular}
\end{table*}

\begin{table*}
\centering
\caption{Comaprison with Schreiber et al.~\cite{schreiber_icdar17} on ICDAR 2013 dataset using the same methods for calculating precision, recall and F1 score as described in Schreiber et al.~\cite{schreiber_icdar17}}
\label{table:schreiber}
\begin{tabular}{| l | r | r |}
    \hline
    & \multicolumn{2}{c}{\textbf{Accuracy\%}} \vline \\\cline{2-3}
    \textbf{Performance Measures} & \textbf{Schreiber et al.} & \textbf{Our Approach} \\ \hline
    Precision & 95.93 & 96.92 \\ \hline
    Recall &  87.36 & 90.12  \\ \hline
    F1 Score & 91.44 & 93.39 \\\hline
    \end{tabular}
\end{table*}

Our proposed solution is also compared with Schreiber et al.~\cite{schreiber_icdar17} which is the state-of-the-art deep learning based approach towards table structure recognition. For that purpose, we used the publicly available ICDAR 2013 table competition dataset containing $67$ documents with $238$ pages, since this dataset was used in~\cite{schreiber_icdar17}. The results of this comparison are shown in Table~\ref{table:schreiber}.

The benchmarking results exhibit that our approach outperforms the existing approaches by a significant margin. There is an increase in the overall correct detections and a decrease in segmentation errors and missed detections. From Table~\ref{column} and~\ref{row}, GRUs outperform LSTMs because of its simpler architecture that is less prone to overfitting.

\section{Conclusion}
This paper proposed a novel approach for table structure extraction using GRU based sequential models for deep learning. This approach provides a significant improvement over heuristic algorithms and CNN based models~\cite{schreiber_icdar17}, owing to the powerful representation of the sequence models that capture the repetitive row/column structures in tables. In the future, we plan to extend this work to develop a coherent framework for information extraction from table cells.

\bibliographystyle{ieeetr}
\bibliography{icdar}

\begin{thebibliography}{10}

\bibitem{lewandowsky}
S.~Lewandowsky and I.~Spence, ``The perception of statistical graphs,'' {\em
  Sociological Methods \& Research}, vol.~18, pp.~200--242, 1989.

\bibitem{schreiber_icdar17}
S.~Schreiber, S.~Agne, I.~Wolf, A.~Dengel, and S.~Ahmed, ``Deepdesrt: Deep
  learning for detection and structure recognition of tables in document
  images,'' in {\em Fourteenth International Conference on Document Analysis
  and Recognition}, vol.~1, pp.~1162--1167, 2017.

\bibitem{unlv}
A.~Shahab, ``Table ground truth for the {UW3} and {UNLV} datasets.''
  \url{http://www.iapr-tc11.org/mediawiki/index.php?title=Table\_
  Ground\_Truth\_for\_the\_UW3\_and\_UNLV\_datasets}, 2010.
\newblock [Online; accessed 7-April-2017].

\bibitem{kieninger_das98}
T.~Kieninger and A.~Dengel, ``A paper-to-html table converting system,'' in
  {\em Proceedings of document analysis systems}, pp.~356--365, 1998.

\bibitem{kieninger_icdar01}
T.~Kieninger and A.~Dengel, ``Applying the {T-RECS} table recognition system to
  the business letter domain,'' in {\em International Conference on Document
  Analysis and Recognition}, p.~0518, 2001.

\bibitem{kieninger99}
T.~Kieninger and A.~Dengel, ``The {T-Recs} table recognition and analysis
  system,'' in {\em Document Analysis Systems: Theory and Practice},
  pp.~255--270, 1999.

\bibitem{kiegner_icapr99}
T.~Kieninger and A.~Dengel, ``Table recognition and labeling using intrinsic
  layout features,'' in {\em International Conference on Advances in Pattern
  Recognition}, pp.~307--316, 1999.

\bibitem{wang_04}
W.~Yalin, I.~T. Phillips, and R.~M. Haralick, ``Table structure understanding
  and its performance evaluation,'' {\em Pattern Recognition}, vol.~37,
  pp.~1479--1497, 2004.

\bibitem{shafait_pami08}
F.~Shafait, D.~Keysers, and T.~M. Breuel, ``Performance evaluation and
  benchmarking of six-page segmentation algorithms,'' {\em IEEE Transactions on
  Pattern Analysis and Machine Intelligence}, vol.~30, no.~6, pp.~941--954,
  2008.

\bibitem{shigarov_16}
A.~Shigarov, A.~Mikhailov, and A.~Altaev, ``Configurable table structure
  recognition in untagged pdf documents,'' in {\em ACM Symposium on Document
  Engineering}, 2016.

\bibitem{zanibbi_das_04}
R.~Zanibbi, D.~Blostein, and R.~Cordy, ``A survey of table recognition: Models,
  observations, transformations, and inferences,'' {\em International Journal
  on Document Analysis and Recognition}, vol.~7, no.~1, pp.~1--16, 2004.

\bibitem{jianying01}
J.~Hu, R.~S. Kashi, D.~P. Lopresti, and G.~Wilfong, ``Table structure
  recognition and its evaluation,'' in {\em Document Recognition and
  Retrieval}, pp.~44--55, 2001.

\bibitem{wangt_dar01}
Y.~Wang, I.~T. Phillips, and R.~Haralick, ``Automatic table ground truth
  generation and a background-analysis-based table structure extraction
  method,'' in {\em Sixth International Conference on Document Analysis and
  Recognition}, pp.~528--532, 2001.

\bibitem{Kasar_icdar_15}
T.~Kasar, T.~K. Bhowmik, and A.~Belaïd, ``Table information extraction and
  structure recognition using query patterns,'' in {\em 13th International
  Conference on Document Analysis and Recognition}, pp.~1086--1090, 2015.

\bibitem{shamilian}
J.~H. Shamilian, H.~S. Baird, and T.~L. Wood, ``A retargetable table reader,''
  in {\em Proceedings of the Fourth International Conference on Document
  Analysis and Recognition}, vol.~1, pp.~158--163, 1997.

\bibitem{siddiqui_decnt}
S.~Siddiqui, M.~I. Malik, S.~Agne, A.~Dengel, and S.~Ahmed, ``Decnt: Deep
  deformable cnn for table detection,'' {\em IEEE Access}, vol.~6,
  pp.~74151--74161, 2018.

\bibitem{gilani_icdar17}
A.~Gilani, S.~R. Qasim, M.~I. Malik, and F.~Shafait, ``Table detection using
  deep learning,'' in {\em 14th International Conference on Document Analysis
  and Recognition}, pp.~771--776, 2017.

\bibitem{shafait_binarize08}
F.~Shafait, D.~Keysers, and T.~M. Breuel, ``Efficient implementation of local
  adaptive thresholding techniques using integral images,'' in {\em SPIE
  Document recognition and retrieval XV}, vol.~6815, p.~681510, 2008.

\bibitem{schuster}
M.~Schuster and K.~K. Paliwal, ``Bidirectional recurrent neural networks,''
  {\em IEEE Transactions on Signal Processing}, vol.~45, no.~11,
  pp.~2673--2681, 1997.

\bibitem{chung14}
J.~Chung, C.~Gulcehre, K.~Cho, and Y.~Bengio, ``Empirical evaluation of gated
  recurrent neural networks on sequence modeling,'' {\em CoRR},
  vol.~abs/1412.3555, 2014.

\bibitem{shahab_das10}
A.~Shahab, F.~Shafait, T.~Kieninger, and A.~Dengel, ``An open approach towards
  the benchmarking of table structure recognition systems,'' in {\em Document
  Analysis Systems}, pp.~113--120, 2010.

\end{thebibliography}
\end{document}